\pdfoutput=1

\documentclass[11pt]{article}

\usepackage[final]{acl}
\usepackage{graphicx}
\usepackage{natbib}
\usepackage{times}
\usepackage{latexsym}
\usepackage[T1]{fontenc}

\usepackage[utf8]{inputenc}

\usepackage{microtype}

\usepackage{graphicx} 
\usepackage{geometry} 
\usepackage{inconsolata}
\usepackage{float}
\usepackage{graphicx}
\usepackage{comment}
\usepackage{todonotes}
\usepackage{url}
\title{Homa at SemEval-2025 Task 5: Aligning Librarian Records with OntoAligner for Subject Tagging}

\author{
 \textbf{Hadi Bayrami Asl Tekanlou\textsuperscript{1}},
 \textbf{Jafar Razmara\textsuperscript{1}},
 \textbf{Mahsa Sanaei\textsuperscript{1}},\\
 \textbf{Mostafa Rahgouy\textsuperscript{2}},
 \textbf{Hamed Babaei Giglou\textsuperscript{3}}
\\
\\
 \textsuperscript{1}University of Tabriz, Tabriz, Iran\\
 \texttt{h.bayrami1403@ms.tabrizu.ac.ir,razmara@tabrizu.ac.ir,mahsa.san75@gmail.com} \\
 \textsuperscript{2}Auburn University, Alabama, USA\\
 \texttt{mzr0108@auburn.edu} \\
 \textsuperscript{3}TIB - Leibniz Information Centre for Science and Technology, Hannover, Germany\\
 \texttt{hamed.babaei@tib.eu}
}

\begin{document}
\maketitle
\begin{abstract}
This paper presents our system, Homa, for SemEval-2025 Task 5: Subject Tagging, which focuses on automatically assigning subject labels to technical records from TIBKAT using the Gemeinsame Normdatei (GND) taxonomy. We leverage OntoAligner, a modular ontology alignment toolkit, to address this task by integrating retrieval-augmented generation (RAG) techniques. Our approach formulates the subject tagging problem as an alignment task, where records are matched to GND categories based on semantic similarity. We evaluate OntoAligner’s adaptability for subject indexing and analyze its effectiveness in handling multilingual records. Experimental results demonstrate the strengths and limitations of this method, highlighting the potential of alignment techniques for improving subject tagging in digital libraries.
\end{abstract}

\section{Introduction}
Libraries are the heart of every society and a cornerstone of education, serving as repositories of human knowledge and cultural heritage. As information landscapes evolve, these institutions must adapt to the growing volume and complexity of digital resources. Therefore, technological innovation in both traditional libraries and modern digital library systems is essential to optimize workflows, enhance accessibility, and improve resource organization. With the rapid advancement of artificial intelligence (AI), particularly through Large Language Models (LLMs)~\cite{chang2024survey}, there is an increasing need to integrate these technologies into library systems~\cite{cox2023chatgpt}. LLMs offer capabilities in natural language understanding (NLU), knowledge retrieval, and automated categorization, making them valuable tools for subject tagging, metadata enrichment, and semantic search~\cite{kasneci2023chatgpt}. By leveraging LLMs, libraries can enhance cataloging efficiency, improve interoperability with controlled vocabularies such as the Gemeinsame Normdatei (GND) ~\cite{GND2025} librarian collections, and enable more precise and context-aware information retrieval.

Despite these advantages, integrating AI-driven solutions into library workflows presents challenges, including model interpretability, bias in automated tagging, and multilingual processing. Addressing these issues requires developing robust frameworks that balance AI-powered automation with human oversight. LLMs4Subjects~\cite{llms4subjects2025} is the first shared task of its kind organized within SemEval-2025, challenging the research community to develop cutting-edge LLM-based solutions for subject tagging of technical records from Leibniz University’s Technical Library (TIBKAT). The participants are tasked with leveraging LLMs to tag technical records using the GND taxonomy. The bilingual nature of the task is designed to address the needs of library systems that often involve multi-lingual records. Given these motivations, the LLMs4Subjects shared task consist of the following two tasks: \textbf{Task 1 -- Learning the GND Taxonomy} -- Incorporating the GND subjects taxonomy, used by Technische Informationsbibliothek (TIB) experts for indexing, into LLMs for subject tagging to enable LLMs to understand and utilize the taxonomy for subject classification effectively. \textbf{Task 2 -- Aligning Subject Tagging to TIBKAT} -- Given a librarian record, a developed system should recommend GND subjects based on semantic relationships in titles and abstracts. 

Ontologies are a key building block for many applications in the semantic web. Hence, ontology alignment, the process of identifying correspondences between entities in different ontologies, is a critical task in knowledge engineering. To this end, OntoAligner~\cite{Babaeillms4om2024,babaei_giglou_ontoaligner_2024} is a comprehensive modular and robust Python toolkit for ontology alignment built to make ontology alignment easy to use for everyone. Inspired by this vision, we adapted its technique for Subject Indexing, where we formulated a dataset into the input data structure of OntoAligner and used the retrieval-augmented generation (RAG) technique to assess the OntoAligner capability in downstream tasks such as subject tagging. The experimental setting in this work plays a case study to analyze OntoAligner behavior toward how much it can be flexible and accurate for subject indexing and what are the bottlenecks.


\section{Related Work}
\label{rel-work}
Subject indexing in library systems has evolved to balance precision and adaptability, incorporating controlled vocabularies, social tagging, ontology-based indexing, and hybrid approaches. Controlled vocabularies, such as the Library of Congress Subject Headings (LCSH), provide structured access to resources but require substantial intellectual effort to maintain consistency \cite{ni2010subject}. While LCSH has expanded through cooperative contributions, it faces criticism for outdated terminology and limited flexibility \cite{pirmann2012tags}. Social tagging, introduced with Web 2.0, allows users to generate metadata, enhancing discoverability and personalization \cite{gerolimos2013tagging, ni2010subject}. However, its effectiveness in library systems remains inconclusive, with studies suggesting that while tags aid browsing, they lack the specificity of controlled vocabularies \cite{rolla2009user, pirmann2012tags}. Ontology-based indexing enhances retrieval accuracy by linking text to structured semantic concepts, addressing limitations of traditional keyword-based indexing \cite{kohler2006ontology}. Hybrid models integrating these approaches are increasingly advocated. Tags can supplement subject headings rather than replace them \cite{gerolimos2013tagging}, as seen in implementations like BiblioCommons \cite{ni2010subject}. However, usability challenges persist, particularly in supporting tag-based searches within catalog interfaces \cite{pirmann2012tags}. This evolving landscape underscores the need for innovative indexing solutions that combine structured control with user-driven flexibility.

\section{Methodology}
\label{method}

In this study, we employ the OntoAligner library~\cite{OntoAligner2025,babaei_giglou_ontoaligner_2024,Babaeillms4om2024} -- a Retrieval-Augmented Generation (RAG) pipeline -- to align technical records from the TIBKAT for Subject Indexing tasks. This task involves generating relevant subject suggestions that accurately reflect the content of a given technical record. The RAG pipeline is designed to handle multilingual, hierarchical data, ensuring that metadata and semantic relationships within the records are preserved for efficient retrieval. Our proposed methodology consists of two main components: 1) OntoAligner Pipeline, and 2) Fine-Tuning.
\subsection{OntoAligner Pipeline}
\noindent\textbf{1) Data Representation.} To align the technical records with the target subjects, we explore multiple levels of information from the records for representation of input data: 1) \textit{Title-based Representation}: We start by using the titles of the technical records, capturing the most concise representation of the content. 2) \textit{Contextual Representation}: We enhance the alignment by incorporating additional metadata, such as abstracts and descriptions, providing deeper context for each record. 3) \textit{Hierarchical Representation}: For records with hierarchical relationships, we include parent-level metadata, enriching the alignment by reflecting the structural relationships within the ontology. These varied representations ensure that both the content and the structural relationships within the records are leveraged to accurately map to relevant subjects.

\noindent \textbf{2) Retrieval Module of OntoAligner.} We employ Nomic-AI embedding models~\cite{nussbaum2024nomic} to generate dense embeddings of the technical records and their corresponding subjects. These embeddings are used to retrieve the top-k most relevant subjects for each record by computing cosine similarity between the record's embedding and the embeddings of potential subjects. We configure the top-k to 30 subject tags.

\noindent \textbf{3) LLM Module of OntoAligner.} The LLM module in OntoAligner leverages advanced language models to enhance the alignment process. This module utilizes Qwen2.5-0.5B~\cite{qwen2.5} to interpret and align complex ontological concepts effectively. By integrating LLMs, OntoAligner can process natural language descriptions and context, facilitating more accurate alignments. After retrieving the top-k relevant candidates for indexing a given librarian record, the LLM evaluates whether each subject is a suitable match or not. This approach follows a RAG paradigm, seamlessly integrating ontology matching within OntoAligner.

\subsection{Fine-Tuning}
Within prior experimentation on three types of input representation -- title, contextual, and Hierarchical -- using the development set and computational resource on hand, we preferred to move forward with \textit{title-based} input representation. In the following, we will discuss the details for retriever and LLM model finetunings.

\begin{table}[t]
    \centering
    \resizebox{0.5\textwidth}{!}{%
    \begin{tabular}{l l c}
         \hline
         \textbf{Sentence 1}& \textbf{Sentence 2} & \textbf{Score} \\
         \hline
         \hline
         Springer eBook Collection & Thermodiffusion & 1 \\
         Springer eBook Collection & Zeitaufl\"osung & 0 \\
         \hline
         ACM Digital Library & Software Engineering & 1\\
         ACM Digital Library & Laser & 0 \\
         \hline
    \end{tabular}}
    \caption{Examples from the retriever model fine-tuning dataset. \textbf{Sentence 1} column represents the title of the librarian record, while \textbf{Sentence 2} column corresponds to the assigned subject. \textbf{Score} column indicates whether the title and subject are a match (1) or not (0).}
    \label{tab:sts}
\end{table}

\noindent\textbf{Contrastive Learning for Retrieval Model. } To fine-tune the retriever module, we constructed a Semantic Textual Similarity (STS)~\cite{majumder2016semantic,giglou2023leveraging} dataset. The records were then paired with their ground truth subjects, assigning a similarity score of 1 for correct pairs. To introduce contrastive learning, we randomly selected negative samples—subjects not associated with the record—and assigned them a similarity score of 0. This resulted in a balanced dataset with \textit{32,952} sentence pairs, ensuring the retriever learns to distinguish relevant subjects from irrelevant ones based on textual similarity. The limit of 600 pairs applied per record from the training set. This threshold is applied to reduce the number of training sets for the retriever module due to the computational resource limitation. The \autoref{tab:sts} represents examples of the obtained datasets for positive and negative pairs. We fine-tuned a sentence-transformer model~\cite{reimers-2019-sentence-bert} (specifically \url{https://huggingface.co/nomic-ai/nomic-embed-text-v1}) using the Multiple Negatives Ranking Loss~\cite{henderson2017efficient}. The model is fine-tuned for 3 epochs with a batch size of 32. The training process leveraged contrastive learning to distinguish between relevant and irrelevant subject pairs, optimizing the model to improve retrieval performance.

\noindent\textbf{Supervised Fine-Tuning of LLM. } We followed a similar process as the retriever model fine-tuning, constructing the fine-tuning dataset with a limit of 200 pairs per record. This resulted in a total of \textit{12,348} samples for supervised fine-tuning (SFT). Later, we fine-tuned a \textit{Qwen2.5-0.5B-Instruct} LLM using QLoRA-based~\cite{dettmers2023qlora} SFT to adapt it for a classification task. The training involved processing the dataset into prompt-based inputs (we used the same as OntoAligner prompts described by \citet{Babaeillms4om2024}), where the model was tasked with determining whether the title and subject tag are match or not. The model was trained over 10 epochs using a batch size of 8, leveraging the Paged AdamW optimizer~\cite{loshchilov2017decoupled} with 8-bit precision for better computational efficiency. The fine-tuned model was then saved for further evaluation using the OntoAligner pipeline.

\section{Results}
\label{results}

\begin{table}[t]
    \centering
    \small
    \begin{tabular}{lccc }
        \hline
        
       \textbf{Dataset} & \textbf{Avg Prec.} & \textbf{Avg Rec.} & \textbf{Avg F1}  \\
        \hline
        \hline
        \multicolumn{4}{l}{\textbf{Quantitative Results}}\\
        \hline
        \textit{TIB-Core}   &  2.84   & 20.30  &  4.66 \\
        \hline
        \multicolumn{4}{l}{\textbf{Qualitative Results}}\\
        \hline
        \textit{Case 1}    & 22.99  & 27.20  & 23.54  \\
        \hline
        \textit{Case 2}    & 14.02  & 23.39  & 16.33   \\
        \hline
    \end{tabular}
    \caption{Quantitative and Qualitative results on TIB-Core-Subjects sets. The averaged metrics are reported.}
    \label{tab:results}
\end{table}

\begin{figure*}[t]
   \centering
   \includegraphics[width=0.8\textwidth]{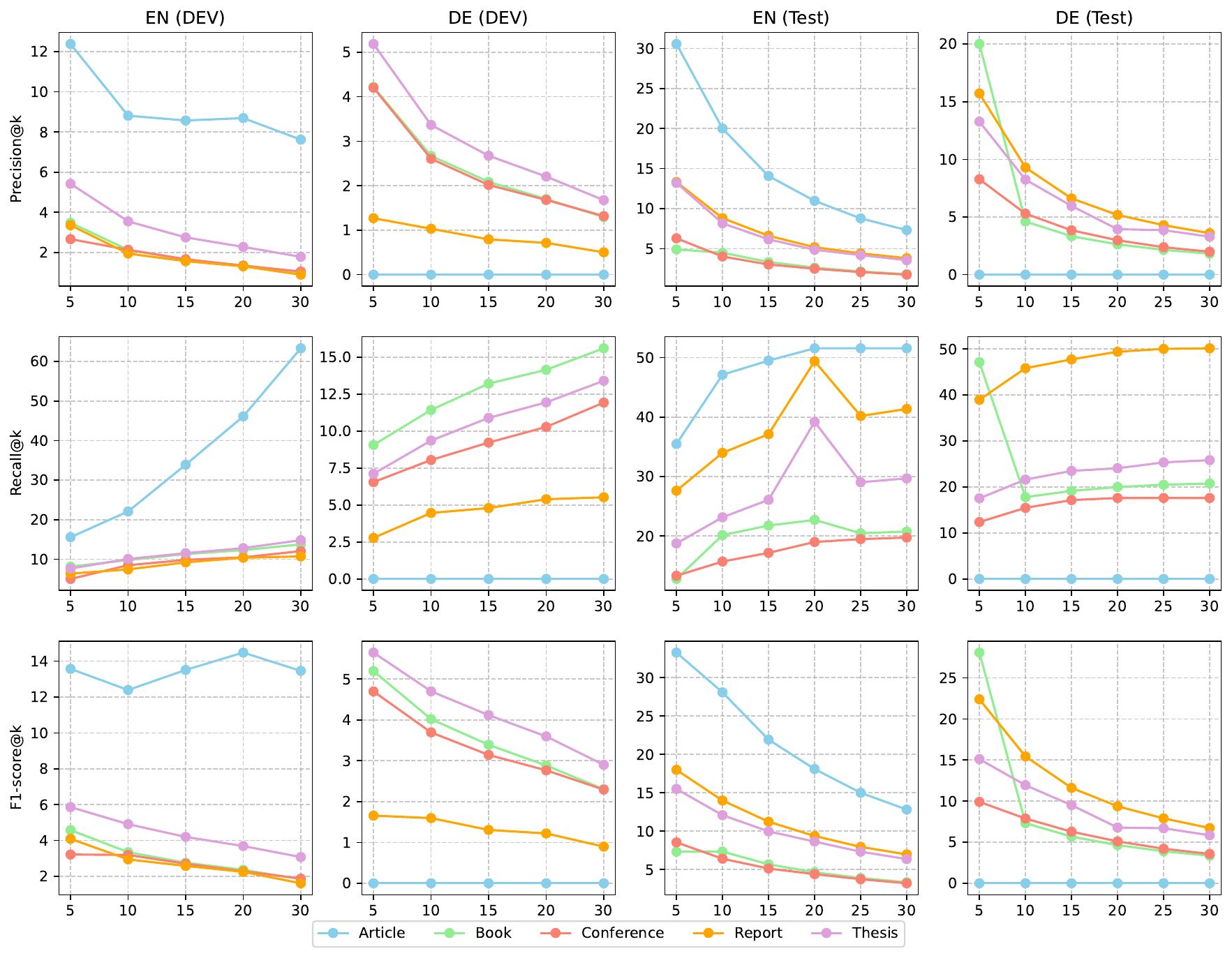}  
   \centering \caption{Results for development and test sets per language and record types.}
   \label{dev-test-linechart}
\end{figure*}

\subsection{Dataset}
For evaluations, we use the TIB-Core-Subjects dataset, which comprises 15,263 technical records across five categories: Article, Book, Conference, Report, and Thesis, in both English and German. Language distribution includes 8,195 English records and 7,113 German records, ensuring a balanced multilingual evaluation. The dataset is split into 7,632 training samples, 3,728 test samples, and 3,948 development samples. 


\begin{figure}[t]
   \centering
   \includegraphics[width=0.5\textwidth]{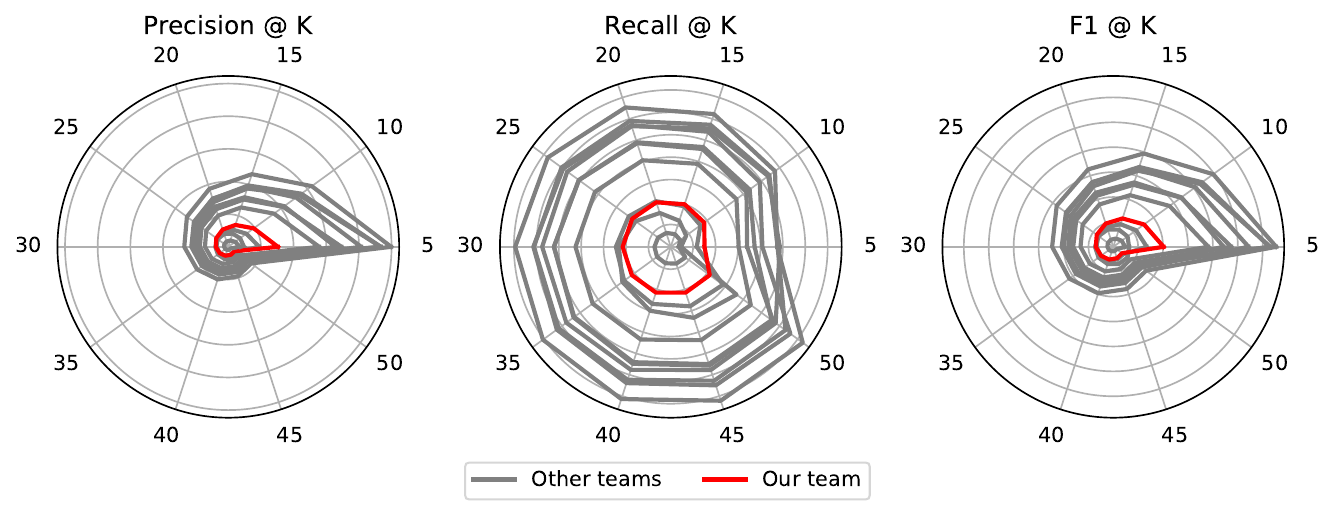}  
   \centering \caption{All the participant results on the test set.}
   \label{test-tibcore-radarplot}
\end{figure}

\subsection{Quantitative Results}

The \autoref{dev-test-linechart} and \autoref{test-tibcore-radarplot} provide a comprehensive comparison of system performance across different languages, record types, and top-k candidates using quantitative metrics. Additionally, \autoref{tab:results} summarizes the average precision, recall, and F1 scores for the quantitative results on the TIB-Core. 

\noindent\textbf{Recall Performance Across k Values.} As we can see within \autoref{dev-test-linechart}, the recall@k curves show a steady increase as k increases, with a notable jump beyond k=15. This pattern suggests that while initial ranked results contain relevant subjects, broader subject coverage improves at higher k values. The German language recall scores remain lower than English, likely due to richer training data or better linguistic resources embedded within LLMs.

\noindent\textbf{Precision Trends Across Languages.} The Precision@k at \autoref{dev-test-linechart} indicate that English consistently outperforms German across both the development and test sets. The English dev and test curves show higher precision values at all k values compared to their German counterparts. This suggests that the subject alignment model is more effective in English, reinforcing the earlier observation of language-based performance differences.

\noindent\textbf{F1 Balance Between Precision and Recall. }F1@k in \autoref{dev-test-linechart} demonstrates a balanced trade-off between precision and recall. The scores peak around k=15–20 before stabilizing, indicating an optimal range where subject retrieval achieves a balance between accuracy and comprehensiveness. Beyond k=20, recall gains do not significantly contribute to F1-score, meaning additional retrieved subjects may include more noise.

\noindent\textbf{Performance Variation by Record Type.} The \autoref{dev-test-linechart} shows that, among record types, \textit{Articles} and \textit{Books} show higher scores across all metrics, suggesting that these records have clearer subject assignments. In contrast, \textit{Conference} and \textit{Reports} records exhibit lower performance, likely due to ambiguous or overlapping subjects. This indicates a need for refined retrieval strategies for these document types and re-checking the ground truths for more clarity.

\begin{figure}[t]
   \centering
   \includegraphics[width=0.5\textwidth]{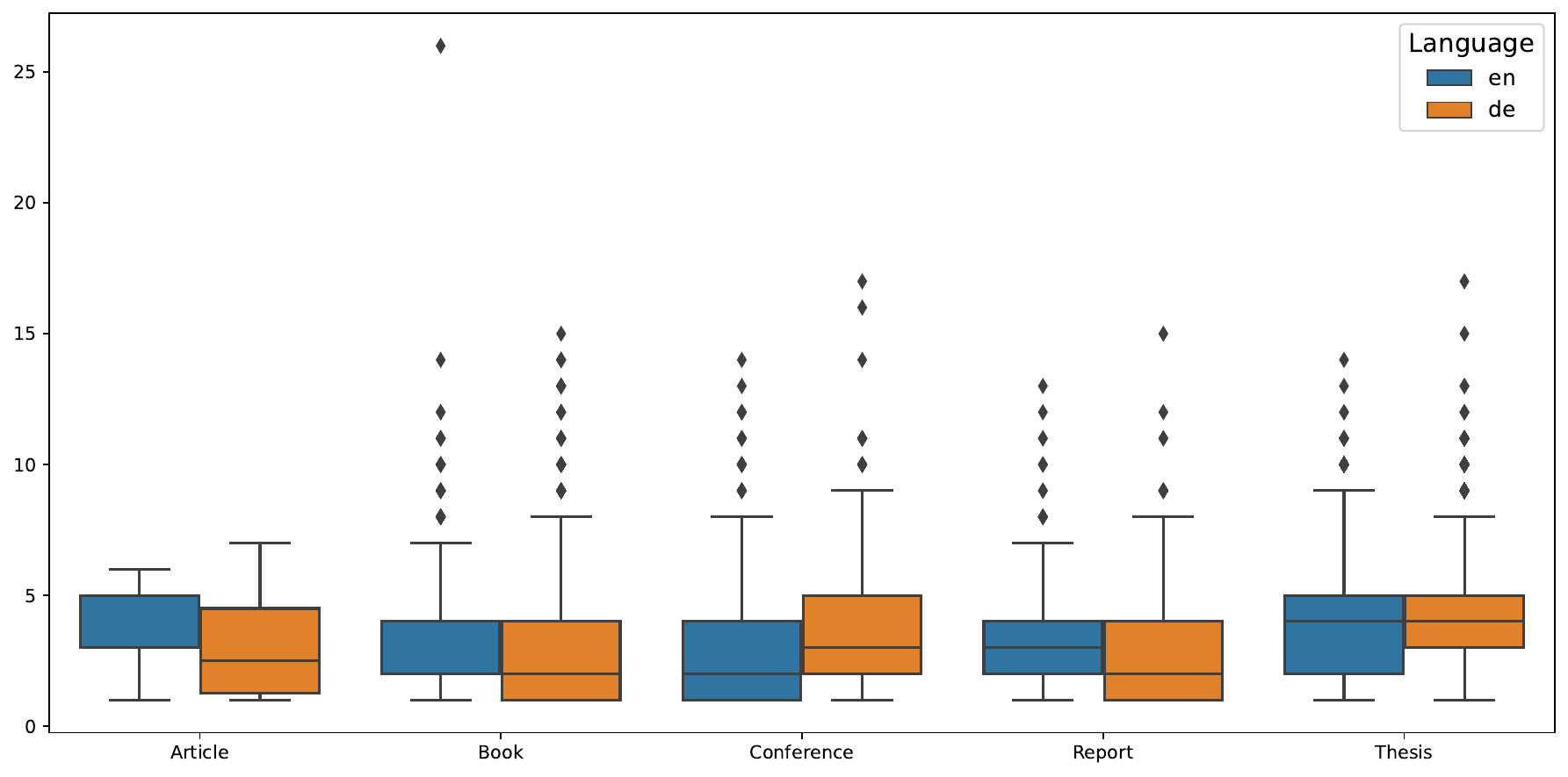}  
   \caption{Distribution of number of subjects across categories using combined train and dev sets}
   \label{tibcore-distribution-boxplot}
\end{figure}

\noindent\textbf{Impact of k Selection on Model Performance.} The choice of k significantly impacts retrieval effectiveness. According to the \autoref{test-tibcore-radarplot} and \autoref{dev-test-linechart}, while lower k values (e.g., k=5) yield higher precision, increasing k enhances recall but at the cost of precision. The optimal balance is observed between k=15 and k=20, where models maintain strong performance without excessive subject list expansion. Furthermore, the distribution analysis of the number of subjects across both languages in \autoref{tibcore-distribution-boxplot} (combination of train and dev sets) indicates that the average number of records typically falls between 0 and 20 with mostly having an upper quartile Q3 of 5. This explains why the results for top-k values within this range vary according to the recall@k in \autoref{test-tibcore-radarplot} for most participants. 

\noindent\textbf{System Performance Against Other Teams.} The \autoref{test-tibcore-radarplot} illustrates our system's performance compared to other teams across different top-k values. While precision differences are marginal, indicating similar ranking effectiveness among top models, the F1 trends show a balance between precision and recall, highlighting our system’s capability in ranking relevant subjects effectively. Additionally, most teams achieved high Recall@5 but lower Precision@5 (with respect to the \autoref{tibcore-distribution-boxplot} this is logical), suggesting that ranking quality is more crucial for retrieval improvements than the LLM module. This is evident in F1@5, where performance drops despite improved recall at k > 5.


\subsection{Qualitative Results}

\begin{figure}[t]
   \centering
   \includegraphics[width=0.5\textwidth, height=4cm]{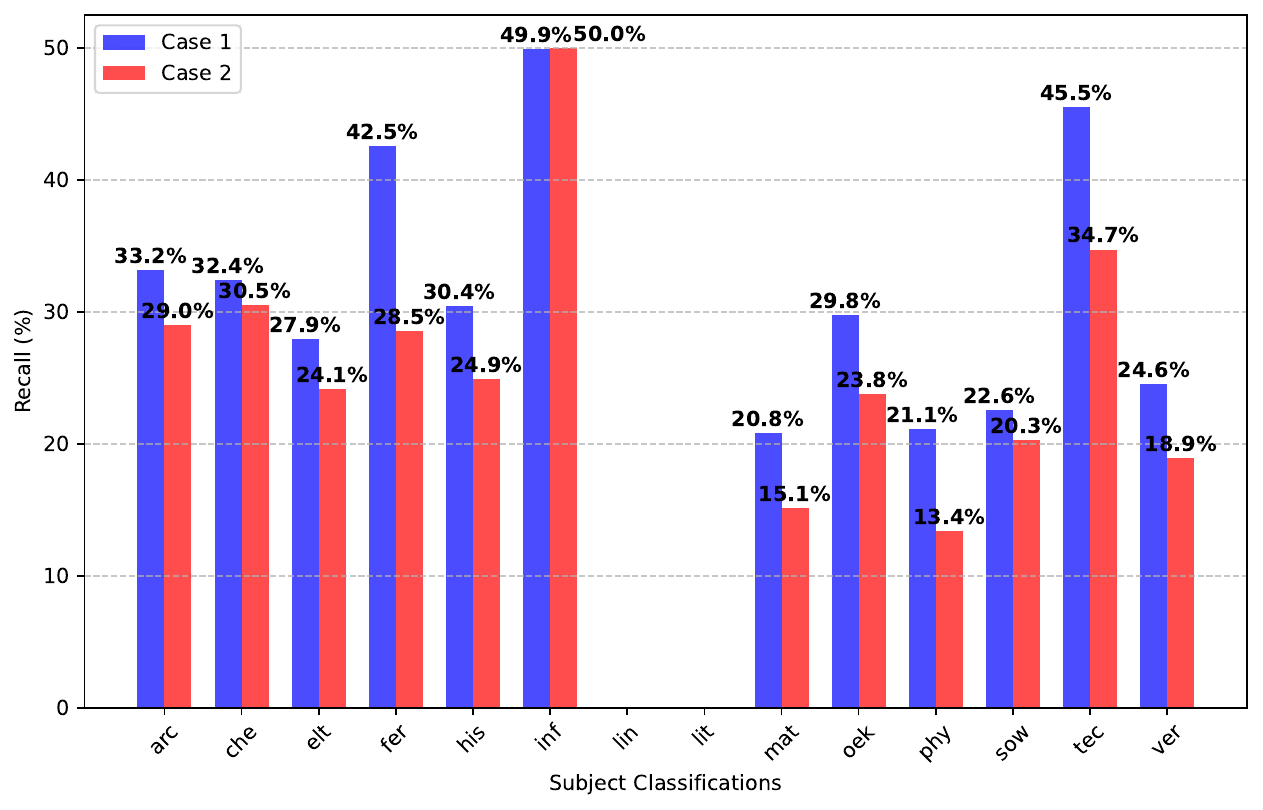}  
    \caption{Recall analysis of qualitative results.}
   \label{test-qual-resuults}
\end{figure}

The \autoref{test-qual-resuults} provides qualitative results for two case studies. Additionally, \autoref{tab:results} summarizes the average precision, recall, and F1 scores for the qualitative results from two case studies.

\noindent\textbf{Case 1 and Case 2 Comparison.} According to the \autoref{tab:results}, the case 1: achieved the highest recall (24.26\%) across all subject classifications, demonstrating that the system effectively retrieves relevant subjects. The F1-score of 20.06\% suggests a balanced trade-off between precision and recall in this scenario, still affected due to the poor precision.  However, case 2 exhibited a lower recall (19.55\%) and F1-score (13.63\%), indicating that the system struggled with certain subject categories, possibly due to more ambiguous or overlapping terms.

\noindent\textbf{Performance Across Subject Classifications.} Figure \ref{test-qual-resuults} further breaks down recall performance by subject classification for both case studies. The highest recall was observed in specific subject categories, such as "inf" (Informatics) -- recall of 50.0\% for case 1 and really of 49.9\% for case 1-- and "tec" (Technology) -- recall of 45.5\% for case 1 and recall of 34.7\% for case 2 --, suggesting that the system performs well in well-structured domains with clear taxonomies. Moreover, the lowest recall of 13.4\% was seen in categories like "phy" (Physics) for case 2 and lowest recall of 20.8\% in "mat" (Mathematics), likely due to their abstract nature and overlapping subject boundaries. Finally, in Case 1, subject categories such as "fer" (Material Science) and "tec" (Technology) performed better compared to Case 2, highlighting the importance of context in subject alignment.

\section{Limitation and Conclusion}
\label{conclusion-findings}
The quantitative evaluation results in \autoref{tab:results} indicate that, despite achieving a strong average recall of 20.30\%, the model struggles with low precision. The low precision suggests that the system retrieves a broad set of candidate subjects, but many are not relevant. However, this is also evident in qualitative results for case-2 where the precision didn't reach the same level as recall. This limitation likely stems from the small fine-tuning dataset, suggesting that further fine-tuning could enhance performance, particularly for smaller LLMs. Additionally, OntoAligner’s flexibility allows rapid pipeline construction by handling embedding storage, subject retrieval, and alignment efficiently. This enables users to focus solely on optimizing the LLM and retriever models, making it practical for subject indexing with minimal resource demands.

In this work, we explored OntoAligner as a case study for subject indexing, demonstrating its capability with minimal fine-tuning. The results highlight its effectiveness in aligning subjects, reinforcing its potential for real-world applications. However, further fine-tuning with additional computational resources and data is necessary to enhance its precision and overall performance for the subject indexing task.

\section*{Acknowledgments}
The last author of this work is supported by the TIB - Leibniz Information Centre for Science and Technology, and the \href{https://scinext-project.github.io/}{SCINEXT project} (BMBF, German Federal Ministry of Education and Research, Grant ID: 01lS22070).

\nocite{*}  
\bibliography{references}

\end{document}